\documentclass[10pt,journal,compsoc]{IEEEtran}

\ifCLASSOPTIONcompsoc
  \usepackage[nocompress]{cite}
\else
  \usepackage{cite}
\fi

\usepackage{graphicx}
\usepackage{ragged2e}
\usepackage{multirow}
\usepackage{amssymb}
\usepackage{amsmath}
\usepackage{booktabs}
\usepackage[switch]{lineno}

\usepackage[ruled,linesnumbered]{algorithm2e}
\usepackage{makecell}  
\usepackage{pifont}
\usepackage{placeins}
\usepackage{selectp}
\SetArgSty{textnormal}
\SetKwInOut{Input}{Input}
\SetKwInOut{Output}{Output}
\DontPrintSemicolon

\makeatletter

\usepackage{makecell}  
\usepackage{amssymb}
\usepackage{pifont}
\usepackage{placeins}
\usepackage{selectp}
\usepackage{times,url,booktabs,microtype,graphicx,bbm,verbatim,amsthm,amssymb,amsmath,enumitem,pifont,makecell,multirow,mdframed,colortbl,rotating,wrapfig,resizegather,resizegather}
\usepackage{subcaption}

\newcommand{\W}{\mathbf{W}}

\newcommand{\X}{\mathbf{X}}
\newcommand{\Y}{\mathbf{Y}}

\newcommand{\B}{\mathbf{B}}

\newcommand{\cmark}{\ding{51}}%
\newcommand{\xmark}{\ding{55}}%
\newcommand{\gr}{\rowcolor[gray]{.95}}
\usepackage{color}
\usepackage{ragged2e}
\usepackage{hyperref}
\usepackage{cite}
\usepackage{array}
\usepackage{colortbl}
\ifCLASSINFOpdf
\else
\fi

\renewcommand{\justify}{\leftskip=0pt \rightskip=0pt plus 0cm}

\begin{document}

\title{Spatial Re-parameterization for N:M Sparsity}

\author{Yuxin Zhang,
        Mingbao Lin,
        Mingliang Xu,~\IEEEmembership{Member,~IEEE}, \\
        Yonghong Tian,~\IEEEmembership{Fellow,~IEEE}, 
        Rongrong Ji,~\IEEEmembership{Senior Member,~IEEE}
\IEEEcompsocitemizethanks{\IEEEcompsocthanksitem Y. Zhang is with the Key Laboratory of Multimedia Trusted Perception and Efficient Computing, Ministry of Education of China, Xiamen University, Xiamen, China.
\IEEEcompsocthanksitem M. Lin is with the Skywork AI, Singapore.
\IEEEcompsocthanksitem M. Xu is with the School of Computer and Artificial Intelligence, Zhengzhou University, Zhengzhou, China, also with Engineering Research Center of Intelligent Swarm Systems, Zhengzhou, China.
\IEEEcompsocthanksitem Y. Tian is with the School of Electronics Engineering and Computer Science, Peking University, Beijing, China.
\IEEEcompsocthanksitem R. Ji (Corresponding  Author) is with the Key Laboratory of Multimedia Trusted Perception and Efficient Computing, Ministry of Education of China, Xiamen University, Xiamen, China, also with Institute of Artificial Intelligence, Xiamen University, Xiamen, China (e-mail: rrji@xmu.edu.cn). 
}
\thanks{Manuscript received April 19, 2005; revised August 26, 2015.}}

\markboth{IEEE TRANSACTIONS ON PATTERN ANALYSIS AND MACHINE INTELLIGENCE}%
{Shell \MakeLowercase{\textit{et al.}}: Bare Demo of IEEEtran.cls for IEEE Journals}

\IEEEtitleabstractindextext{%
\begin{abstract}
\justify{This paper presents a Spatial Re-parameterization (SpRe) method for the N:M sparsity.
SpRe stems from an observation regarding the restricted variety in spatial sparsity of convolution kernels presented in N:M sparsity compared with unstructured sparsity.
Particularly, N:M sparsity exhibits a fixed sparsity rate within the spatial domains due to its distinctive pattern that mandates N non-zero components among M successive weights in the input channel dimension of convolution filters.
On the contrary, we observe that conventional unstructured sparsity displays a substantial divergence in sparsity across the spatial domains, which we experimentally verify to be very crucial for its robust performance retention compared with N:M sparsity.
Therefore, SpRe employs the spatial-sparsity distribution of unstructured sparsity by assigning an extra branch in conjunction with the original N:M branch at training time, which allows the N:M sparse network to sustain a similar distribution of spatial sparsity with unstructured sparsity.
During inference, the extra branch can be further re-parameterized into the main N:M branch, without exerting any distortion on the sparse pattern or additional computation costs.
SpRe has achieved a commendable feat by matching the performance of N:M sparsity methods with state-of-the-art unstructured sparsity methods across various benchmarks.
Our project is available at \url{https://github.com/zyxxmu/SpRE}.
}
\end{abstract}

\begin{IEEEkeywords}
Convolutional neural networks, Network sparsity, N:M sparsity.
\end{IEEEkeywords}}

\maketitle

\IEEEpeerreviewmaketitle

\section{Introduction}\label{introduction}

\IEEEPARstart{N}{etwork} sparsity has proven many successes in reducing the complexity of convolutional neural networks (CNNs)~\cite{han2015learning, lecun1989optimal, luo2017thinet}.
%
A sparse network can be obtained by zeroizing weights at different granularity levels, from fine to coarse.
Fine-grained sparsity (unstructured sparsity)~\cite{lecun1989optimal, ding2019global} removes individual weights and is demonstrated to well retain performance even at high sparsity rates. 
Unfortunately, the deployment of fine-grained sparse networks on off-the-shelf hardware is cumbersome due to the irregularity of sparse weight matrices.
On the other hand, coarse-grained sparsity (structured sparsity)~\cite{he2017channel,lin2020hrank} achieves significant acceleration by eliminating entire convolution filters~\cite{liu2019metapruning, lin2020hrank} or weight blocks~\cite{ji2018tetris, meng2020pruning}, yet suffers severe performance degradation particularly at high sparsity rates.
%


\begin{figure*}[!t]
\begin{center}
\includegraphics[height=0.43\linewidth]{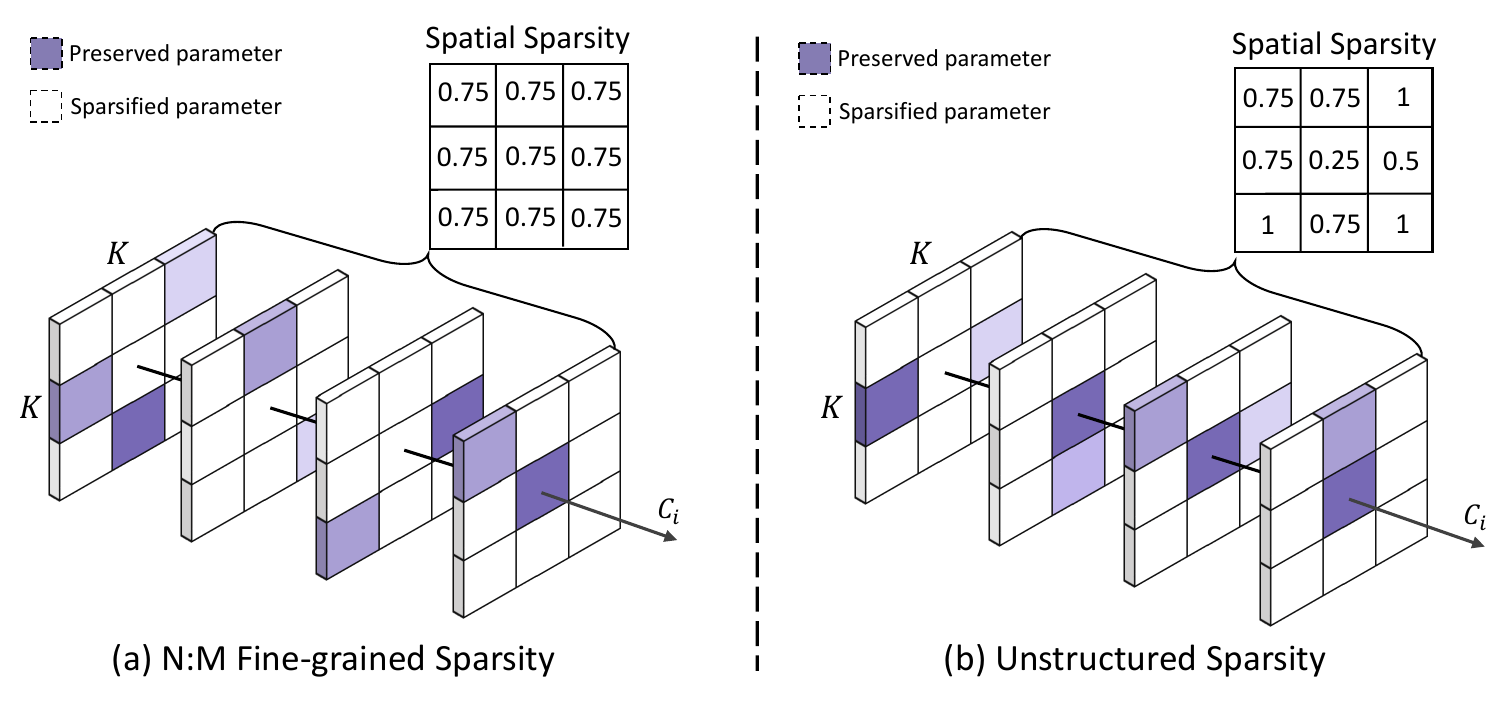}
\vspace{-0.3cm}
\end{center}
\caption{A toy example of the discrepancy in spatial sparsity between N:M sparsity at 1:4 pattern and 75\% unstructured sparsity. {\color{black}We define spatial sparsity as the sparse level across each spatial location of convolution filters,~\emph{i.e.}, the input channel dimension of weight matrices $C_{i}$. (a) N:M sparsity requires N non-zero components among M consecutive weights in the input channel dimension, resulting in equal spatial sparsity $((1-M)/N)$. (b) Unstructured sparsity removes weights at arbitrary locations, resulting in uneven spatial sparsity.}\label{fig:toy_example}}
\vspace{-0.2cm}
\end{figure*}

%
N:M sparsity has lately surfaced as a promising direction of augmenting the trade-off between acceleration effects and performance retention~\cite{zhou2021learning, pool2021channel}.
By stipulating at most N non-zero components within M consecutive weights across the input channel dimension, it considerably improves the performance of structured sparsity while simultaneously ensuring expeditious inference aided by the N:M sparse tensor core~\cite{nvidia2020a100}.
In recent years, various methods have been proposed to train N:M sparse networks from pre-trained weights~\cite{nvidia2020a100, pool2021channel} or randomly-initialized weights~\cite{zhou2021learning,zhang2022learning}.
Nevertheless, the performance of N:M sparsity still lags behind unstructured sparsity, particularly at high sparsity rates such as 95\%~\cite{liu2021sparse, zhou2021learning}.
We ask: \textit{what leads to the performance gap
between N:M sparsity and unstructured sparsity?}

In this paper, we answer this question through empirical observation of network sparsity over the spatial domain.
Particularly, N:M sparsity displays a consistent sparsity rate of $1-\frac{\text{N}}{\text{M}}$ at every spatial location of convolution filters owing to the distinctive sparse pattern depicted in the input channel dimension (Figure\,\ref{fig:toy_example}a).
Conversely, unstructured sparsity can exhibit a notable variation in spatial sparsity (Figure\,\ref{fig:toy_example}b), which we confirm to be ubiquitous in existing unstructured sparsity methods and crucial for their robust performance retention compared with N:M sparsity methods (Sec.~\ref{sec:spatial_vari}).
%
%
{\color{black}To explain this, a heterogeneous distribution of weights across various positions  facilitates prioritization of salient visual features within each local receptive field during sliding window operations, thereby yielding superior performance of sparse networks.
For instance, weights that are located in the center of local receptive field exhibit more significant importance for the shallow network layers, as shown in Figure\,\ref{variant_spatial}.
}
%
Self-evidently, N:M sparsity falters to assign adequate weights for the informative visual elements especially when the sparsity rate is high, therefore leading to more performance degradation.

Driven by this analysis, we present Spatial Re-parameterization (SpRe) as a way of matching the performance between N:M sparsity and unstructured sparsity.
SpRe utilizes the spatial sparsity distribution of unstructured sparsity to allocate an extra weight branch in conjunction with the original N:M branch.
This enables the sparse network unifying these two branches to maintain a sparsity distribution comparable to that of unstructured sparsity.
Moreover, we constrain the newly introduced parameters to adhere to the N:M sparse distribution of the main branch in the input channel dimension.
This results in an advantage for a re-parameterization after training, where the newly-added branch can be merged into the main block without impacting the output at the inference stage.
Thus, SpRe introduces no additional inference burden for the original N:M sparse networks.
The advantages of our proposed SpRe include:
\begin{itemize}
    \item Trackable. SpRE is traceable in principle, due to our innovative observation of the discrepancy in spatial sparsity between N:M sparsity and unstructured sparsity, which we prove to be essential for performance retention.
    \item Scalable. SpRe is easy to use and orthogonal to enhance the effectiveness of other N:M sparsity methods, whether applied from randomly-initialized or pre-trained weights.
    \item High-performance. SpRe is validated to be highly successful in boosting the performance of N:M sparsity methods across various benchmarks. Specifically, SpRe enhances the Top-1 accuracy of SR-STE~\cite{zhou2021learning}, a leading N:M method, by 1.2\% when training a 1:16 sparse ResNet-50~\cite{he2016deep} on ImageNet~\cite{deng2009imagenet}. Moreover, the boosted performance even surpasses state-of-the-art unstructured sparsity method GraNet~\cite{liu2021sparse} by 0.4\% at a similar sparsity rate.
\end{itemize}

\section{Related Work}
\subsection{Unstructured Sparsity}
By removing individual weights at arbitrary positions of the network to address the heavy parameter burden, unstructured sparsity has emerged as a fervent area of research over the last decade~\cite{lecun1989optimal, han2015learning, louizos2017learning}.
Gradient~\cite{lee2018snip}, momentum~\cite{ding2019global}, magnitude~\cite{han2015learning}, and many others, are often used to identify and remove insignificant weights.
Recent advancements learn to train an unstructured sparse network in a dynamic sparse manner towards performance enhancement.
RigL~\cite{evci2020rigging} alternatively removes and revives weights based on magnitudes and gradients.
Sparse Momentum~\cite{dettmers2019sparse} considers the mean momentum magnitude in each layer to redistribute weights.
Besides, gradual sparsity is widely adopted to boost performance~\cite{zhu2017prune,liu2021sparse}.
Unstructured sparsity is demonstrated to well retain performance, even at a very high sparsity over 95\%~\cite{liu2021sparse}. However, the resulting irregular sparse tensors make it gain rare speedups on general hardware~\cite{wang2020sparse}.
%

\begin{figure*}[!t]
\begin{center}
\includegraphics[height=0.4\linewidth]{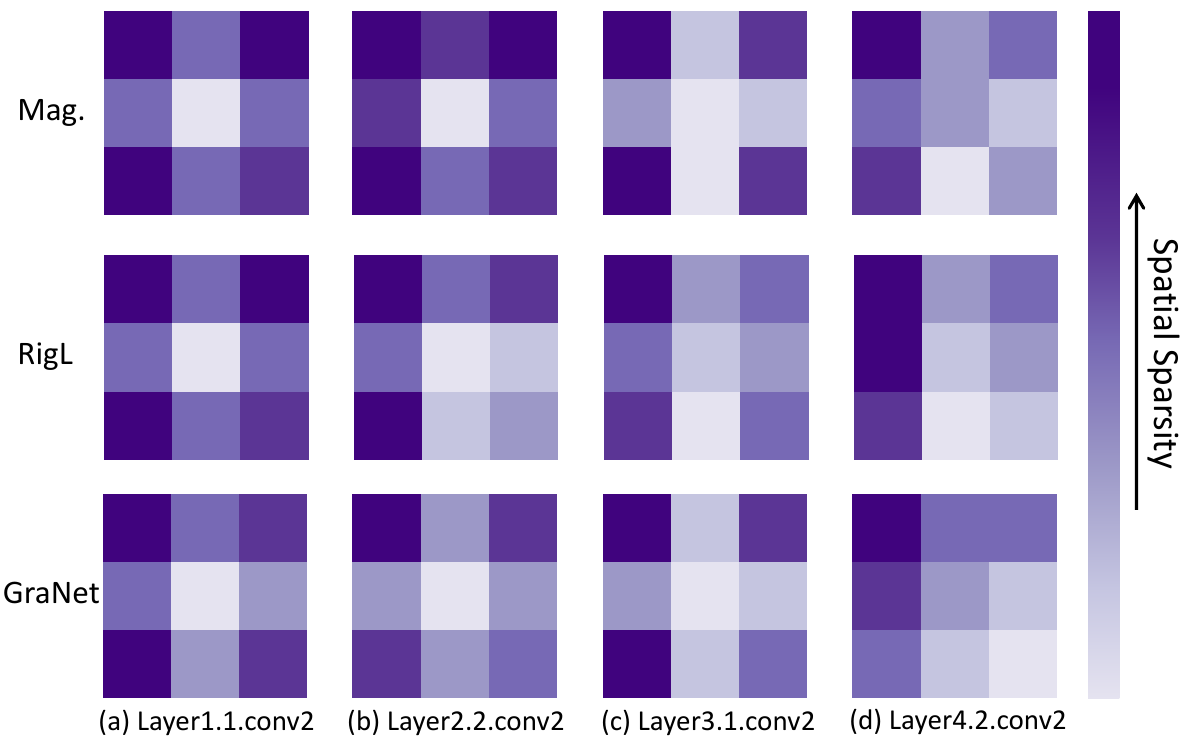}
\end{center}
\vspace{-0.3cm}
\caption{
Spatial sparsity of common unstructured sparsity methods including Magnitude-based sparsity~\cite{han2015learning}, RigL~\cite{evci2020rigging}, GraNet~\cite{liu2021sparse}. We show spatial sparsity of 3$\times$3 kernels from different layers of ResNet-50~\cite{he2016deep} with overall 95\% sparsity, close to the sparsity level of 1:16 pattern. Experiments are performed on ImageNet-1K~\cite{deng2009imagenet}.
}\label{variant_spatial}
\vspace{-0.2cm}
\end{figure*}

\subsection{N:M Sparsity}
N:M sparsity preserves N out of M consecutive weights along the input channel dimension of CNNs, and achieves practical speedups thanks to the hardware innovation of N:M sparse tensor core~\cite{nvidia2020a100, fang2022algorithm}.
The pioneering ASP~\cite{nvidia2020a100} goes through model pre-training, high-magnitude weight removal~\cite{han2015learning}, and model fine-tuning.
Pool~\emph{et al.}~\cite{pool2021channel} proposed channel permutation to increase the performance of ASP.
Sun~\emph{et al.}~\cite{sun2021dominosearch} proposed a layerwise fine-grained N:M sparsity to replace the common uniform version.
To avoid heavy burden on model pre-training, Zhou~\emph{et al.}~\cite{zhou2021learning} proposed a sparse-refined straight-through estimator (SR-STE) to learn from scratch.
More specifically, N-out-of-M weights of higher magnitudes are selected in each forward pass, whileall weights are updated during the backward phase, utilizing the STE estimator, paired with a uniquely designed sparse penalty term.
LBC~\cite{zhang2022learning} reformulates the N:M sparsity as a combinatorial problem and learns the best combination for the sparse weights.
Recent advances~\cite{zhang2023bi, hubara2021accelerated} also excavate into the training efficiency of N:M sparsity through devising masks that enable N:M sparse computation in both forward and backward propagation.

\subsection{Structural Re-parameterization}
Structural re-parameterization mutually converts different architectures through an equivalent transformation of parameters.
The representative RepVGG~\cite{ding2021repvgg} merges kernels of smaller sizes to these of larger ones in inference.
For example, 1$\times$1 kernels can be added onto the central points of the 3×3 kernels.
Along this line, researchers have devised various blocks to boost the performance of a regular CNN without extra inference costs,~\emph{e.g.}, Asymmetric Convolution Block (ACB) ~\cite{ding2019acnet}, RepLKNet~\cite{ding2022scaling}.
Besides, structural re-parameterization is also leveraged to guide channel pruning~\cite{ding2021resrep}, where the original CNN is re-parameterized into two parts to respectively maintain the performance and prune convolutional filters.

In this paper, we focus on developing an N:M sparsity method that is orthogonal to the aforementioned methods for performance improvement, simply due to two advantages of our method:
First, we utilize the distribution of unstructured sparsity across the spatial domain.
Second, we re-parameterize a newly-added branch into the main branch of N:M block in the inference.

\section{Methodology}\label{methodology}

\subsection{Background}

For convolutional weights $\W \in \mathbb{R}^{C_o \times C_i \times K \times K}$ ($C_o$: output channel, $C_i$: input channel, $K$: kernel size), the convolution operation with input features $\X \in \mathbb{R}^{H \times W \times C_i}$ ($H$: height, $W$: width) is generally formulated as:
\begin{equation}\label{eq:forward}
    \Y = BN(\W \circledast \X),
\end{equation}
where $\circledast$ represents convolution operation and $BN(\cdot)$ stands for the follow-up batch normalization~\cite{ioffe2015batch}.
Network sparsity can be realized with a $0$-$1$ mask $\B$ of the same shape to $\W$:
\begin{equation}\label{eq:sparse_forward}
    \Y = BN( (\B \odot \W) \circledast \X),
\end{equation}
where $\odot$ denotes the element-wise multiplication. It is easy to know that, $\B_{p,q,u,v}$ = $0$ removes $\W_{p,q,u,v}$ and $\B_{p,q,u,v}$ = $1$ preserves $\W_{p,q,u,v}$.

For unstructured sparsity, the zero entries are irregular in the positions of $\B$. 
Instead, N:M sparsity stipulates N non-zero entries for every M consecutive weights along the input channel dimension. Therefore, $\B$ is restricted to satisfy:
\begin{equation}\label{eq:nm_constraint}
{\Vert\B_{p, \lfloor q/ \text{M}\rfloor\cdot\text{M}:\lfloor q/ \text{M}\rfloor\cdot\text{M}+\text{M}, u, v}\Vert}_0 = \text{N},
\end{equation}
where $p,q,u,v$ enumerates $C_o, C_i, K, K$, respectively.
Besides, for ease of the following representation, we define Spatial Sparsity (SS) to measure the weight sparsity in the spatial domain. The spatial sparsity at position ($u$, $v$) is calculated as:
\begin{equation}\label{eq:spatial_sparsity}
\text{SS}(\B, u, v) = 1 - \frac{1}{C_o \cdot C_i}\sum_{p =1}^{C_o} \sum_{q =1}^{C_i} \B_{p, q, u, v}.
\end{equation}

\subsection{Discrepancy in Spatial Sparsity}\label{sec:spatial_vari}
Compared to unstructured sparsity, N:M sparsity achieves practical speedups with the support of the N:M sparse tensor core~\cite{nvidia2020a100, fang2022algorithm}, yet it suffers more performance degradation, particularly at high sparsity rates~\cite{zhou2021learning, liu2021sparse}.
In this section, we demonstrate that such a performance gap stems from a discrepancy in the spatial sparsity between N:M sparsity and unstructured sparsity.
Simply due to the constraint of Eq.\,(\ref{eq:nm_constraint}), N:M sparsity possesses consistent spatial sparsity across different positions as:
\begin{equation}\label{eq:spatial_nm}
    \text{SS}(\mathbf{B}, u, v) = 1 - \frac{\text{N}}{\text{M}}.
\end{equation}

In other words, N:M sparsity equally allocates weights across the spatial domain of convolution operation.
On the other hand, such balanced spatial sparsity is not imperative for unstructured sparsity, given that no sparsity restriction is imposed across the channel dimension.
Indeed, we observe that unstructured sparsity exhibits significant variability in terms of spatial sparsity, as illustrated by Figure\,\ref{variant_spatial}.
%
This variability in spatial sparsity remains consistent across different layers and network types, irrespective of the specific unstructured sparsity technique employed~\cite{han2015learning, evci2020rigging, molchanov2016pruning}.
%
%
%

Upon closer examination of Figure\,\ref{variant_spatial}, it can be inferred that the majority of unstructured weights even maintain a similar distribution spatial sparsity, more interestingly resembling a cross-like configuration in the shallow layers.
This intriguing phenomenon warrants further exploration and investigation within the domain of unstructured sparsity.
We do not delve deeply into this matter at present, but it is unequivocal that unstructured sparsity methods~\cite{han2015learning, evci2020rigging, molchanov2016pruning} allocate more weights to certain fixed visual points simultaneously. 

\begin{figure*}[!t]
\begin{center}
\includegraphics[height=0.4\linewidth]{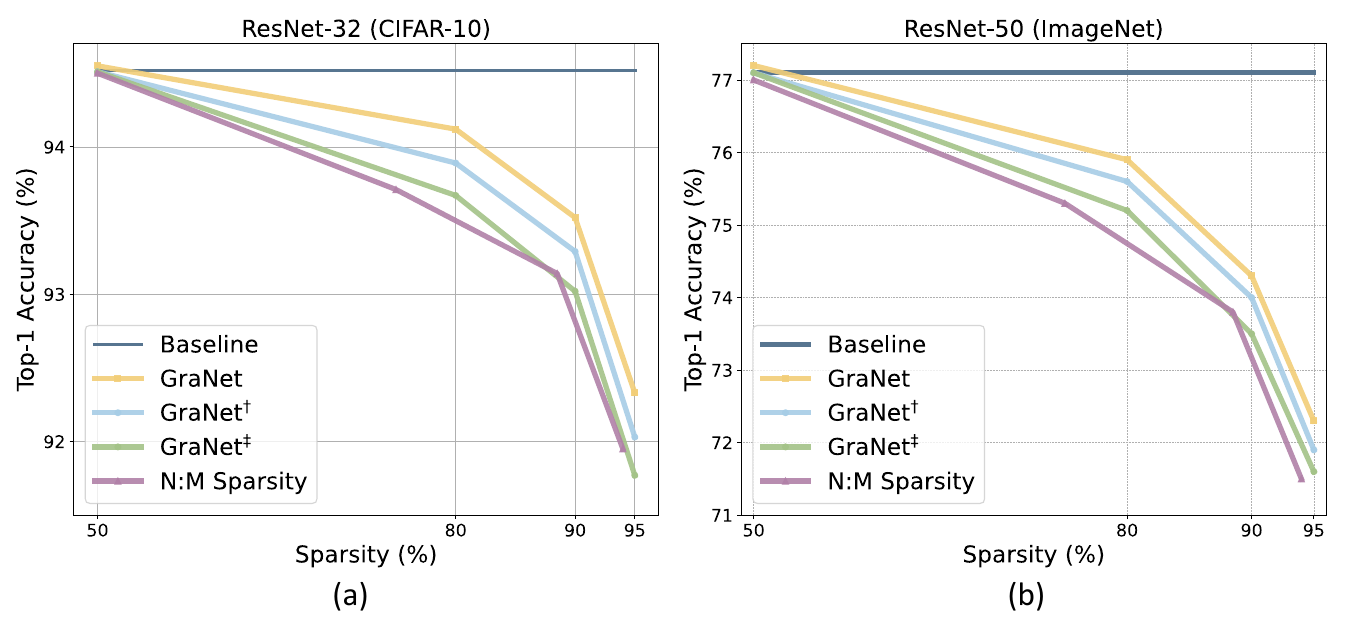}
\end{center}
\vspace{-0.4cm}
\caption{
Performance comparison for pruning ResNet-32~\cite{he2016deep} on CIFAR-10~\cite{krizhevsky2009learning} and ResNet-50~\cite{he2016deep} on ImageNet-1K~\cite{deng2009imagenet}. The involved methods include the dense version of ResNet-50 (Baseline), SR-STE~\cite{zhou2021learning} (N:M sparsity), unstructured sparse GraNet~\cite{liu2021sparse}, and GraNet with two types of sparsity constraints: (1) The same spatial sparsity along the spatial dimension (GraNet$^{\dag}$), (2) Equal mask flexibility with (1) but no spatial sparsity constraint (GraNet$^{\ddag}$).
Performance drops when unstructured sparse method is confined to having the same spatial sparsity, even if the mask flexibility remains the same.
}\label{fig:even_spatial_performance}
\vspace{-0.2cm}
\end{figure*}

CNNs analyze visual imagery by sliding filters along input features to empower each weight to interact with different visual regions. 
{\color{black}We conjecture that unstructured sparsity methods inherently acquire weights in the spatial domain that can better prioritize salient visual features within each local receptive field during sliding window operations, thereby enhancing the performance of sparse networks.

As illustrated in Figure\,\ref{fig:even_spatial_performance}, weights positioned at the center of the local receptive field demonstrate greater significance for shallow network layers those primarily capture low-level texture information. In contrast, deeper layers of the network are tasked with extracting more abstract semantic features, leading to a seemingly random distribution of important weights,~\emph{i.e.}, spatial sparsity distribution across spatial dimensions.
}
To verify our hypothesis, we destroy the variable spatial sparsity of the state-of-the-art unstructured sparsity method GraNet~\cite{liu2021sparse} by constraining the sparsity at any location ($u$, $v$) to be the same.
In particular, given a target sparsity rate $P$, we restrict the binary mask $\B$ to satisfy:
\begin{equation}\label{eq:unstructure_spatial_constraint}
{\Vert\B_{:,:,u,v}\Vert}_0 \le \lfloor {P \cdot \Vert\W_{:,:,u,v}\Vert}_0 \rfloor.
\end{equation}

The above constraint leads to the same spatial sparsity at any location, as well as limiting the flexibility of pruning masks, which refers to the solution space for network sparsity and is a key factor for performance preservation~\cite{hoefler2021sparsity, lin20221xn}.
For a more rigorous comparison, we provide another sparsity constraint by firstly reshaping $\W \in \mathbb{R}^{C_{o} \times C_{i}  \times  K \times K }$ into $\W' \in \mathbb{R}^{ K \times K \times C_{o} \times C_{i} }$, then the binary mask $\B$ is restricted as:
\begin{equation}\label{eq:alternative}
{\Vert\B_{p,q,:,:}\Vert}_0 \le \lfloor {P \cdot \Vert\W'_{p,q,:,:}\Vert}_0 \rfloor.
\end{equation}

Eq.\,(\ref{eq:unstructure_spatial_constraint}) and Eq.\,(\ref{eq:alternative}) have the same mask flexibility, except that the former enforces sparsity in the spatial domain.
Then, we perform experimental comparison in Figure\,\ref{fig:even_spatial_performance}.
Compared to the vanilla unstructured method~\cite{han2015learning}, limited sparsity flexibility incurs performance degradation, even close to N:M method SR-STE~\cite{zhou2021learning}.
Furthermore, requirement of spatial sparsity by Eq.\,(\ref{eq:unstructure_spatial_constraint}) is not a patch on that by Eq.\,(\ref{eq:alternative}).
To explain, when weights are scarce at elevated sparsity rates, it becomes more imperative to direct the constrained weights towards more critical visual points in order to guarantee performance retention.
Conversely, if spatial sparsity is maintained at an equilibrium level, these vital visual points cannot be efficiently processed, thereby leading to more performance degradation.
Therefore, we can ascertain the core role of spatial sparsity variability in improving the performance of N:M methods.

\begin{figure*}[t]
\begin{center}
\includegraphics[height=0.5\linewidth]{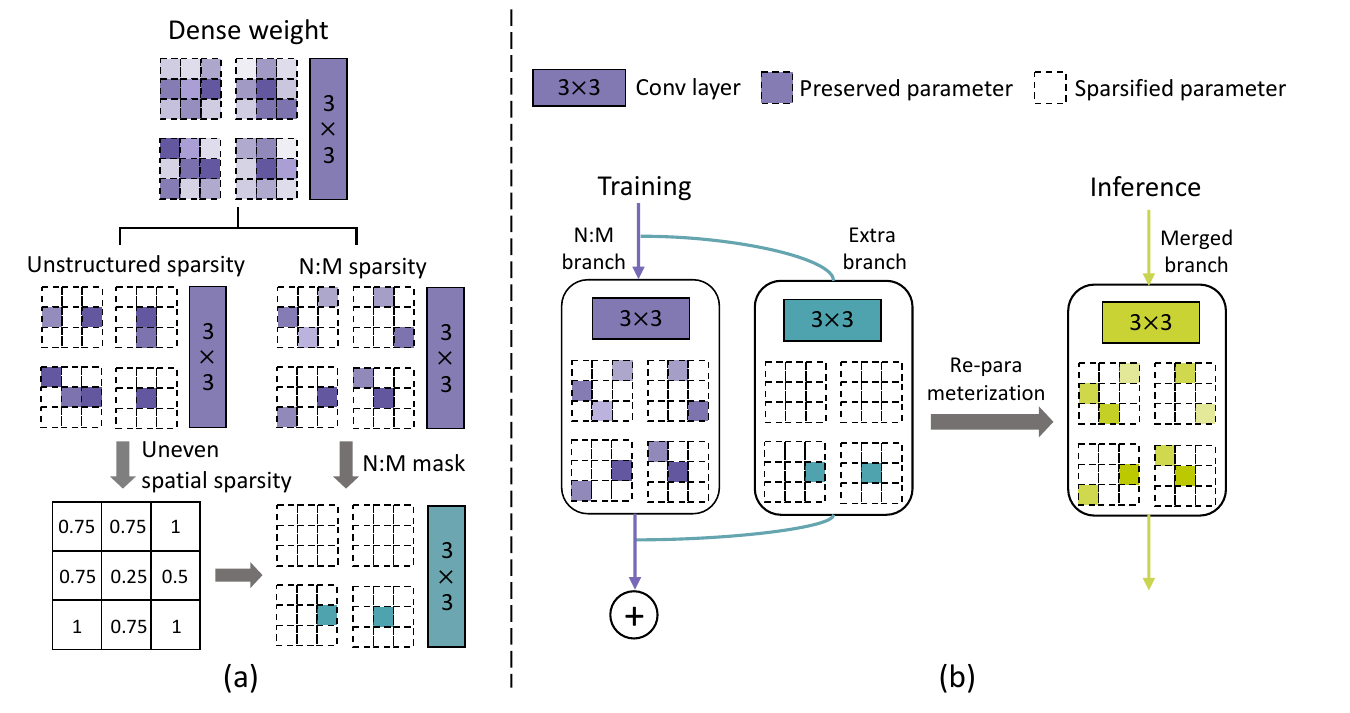}
\end{center}
\vspace{-0.3cm}
\caption{Framework of SpRe. {\color{black}(a) The uneven spatial sparsity arising from unstructured sparse weights is leveraged to build an extra branch that reimburses the spatial sparsity of N:M mask (Eq.\,(\ref{eq:spatial_mask})). (b) The extra branch is then trained in conjunction with the main N:M branch to mitigate the performance drop caused by spatial sparsity gap between unstructured sparsity and N:M sparsity. After training, a re-parameterization is performed to merge these two branches to get an N:M sparse merged branch, without altering the output.}
}\label{fig:method}
\vspace{-0.2cm}
\end{figure*}

\subsection{Spatial Re-parameterization}
We propose Spatial Re-parameterization (SpRe), as depicted in Figure\,\ref{fig:method}, as a way of attaining comparable performance in N:M sparsity to that of unstructured sparsity.
{\color{black}To address the issue of limited spatial sparsity variation in N:M, SpRe introduces an extra branch in conjunction with the original N:M sparse weights $\B \odot \W$.
The weights of this new branch are denoted as $\W^S \in \mathbb{C}_o \times \mathbb{C}_i \times K \times K$, with a corresponding mask $\B^S$ of the same shape.}
Specifically, we first obtain mask $\B^U$ of unstructured sparsity with a sparsity rate $P=1-\frac{\text{N}}{\text{M}}$.
Here we utilize the classic magnitude-based pruning~\cite{han2015learning}, while other metrics for unstructured sparsity~\cite{molchanov2016pruning, lee2018snip, wang2020picking} can also adapt well.
Given that N:M weights $\B \odot \W$ maintain a constant spatial sparsity of rate $1-\frac{\text{N}}{\text{M}}$, we allocate less parameter burden in the extra unstructured branch as:
\begin{equation}\label{eq:spatial_mask}
\B^S_{:,:,u,v} = \left\{ \begin{array}{ll} 
 \B_{:,:,u,v}, \; \textrm{if} \; \text{SS}(\B^U, u,v) < 1- \frac{\text{N}}{\text{M}}, \\
\mathbf{0}, \; \textrm{otherwise,}
  \end{array} \right.
\end{equation}
in which $u, v = 1, ..., K$.
Then, the regular forward propagation changes from Eq.\,(\ref{eq:sparse_forward}) to:
\begin{equation}\label{eq:forward_train}
    \Y = BN((\B \odot \W) \circledast \X) + BN(( \B^S \odot \W^S) \circledast \X).
\end{equation}

Consequently, the total spatial sparsity, unifying these two branches, is equivalently compensated to a correspondingly leveled unstructured sparsity during the training phase.
Such a practice allows the sparse network to emphasize crucial visual regions, thereby yielding enhanced {\color{black}performance.
Next}, we demonstrate that the extra branch can be re-parameterized into the main branch without incurring additional parameters or computation burden at the test stage. 
Particularly, in each branch, the BN layer and its preceding convolution layer can be converted into a single convolution kernel, which is a very widely-used technique in the deployment of CNN networks~\cite{ding2021repvgg}.
Denoting the converted weight in the main and extra branches as $\B \odot \bar{\W}$ and $\B^S \odot \bar{\W}^S$, the final conv kernel $\W^*$ is derived by point-wise addition of the converted weights as:
\begin{equation}\label{eq:merge}
    \W^* = \B \odot \bar{\W} + \B^S \odot \bar{\W}^S.
\end{equation}

Given that the mask of the extra branch is a subset of the main branch ($\B^S$ is a subset of $\B$), as per Eq.\,(\ref{eq:spatial_mask}), the merged weights are still in the N:M form for inference. 
Eventually, the forward propagation at the test stage is conducted with N:M sparse weights as:
\begin{equation}\label{eq:forward_inference}
    \Y =  \W^* \circledast \X.
\end{equation}
Self-evidently, neither interference on the N:M pattern nor extra inference burden is introduced by SpRe.
\textbf{Implementation of SpRe.} SpRe is orthogonal to existing N:M sparsity techniques and can be seamlessly combined to enhance their lack of spatial sparsity variety. 
For techniques that attain N:M sparsity after pre-training the dense weights~\cite{han2015learning, pool2021channel}, SpRe can be directly applied to build the re-parameterization branch on the basis of pre-trained weights.
After fine-tuning, we employ the re-parameterization to derive the N:M sparse weights for inference.
For methods that excavate N:M sparsity from scratch in a sparse training manner~\cite{zhou2021learning, zhang2022learning}, SpRe can also dynamically adapt the binary mask $\B^S$ of the extra branch by looking at the updated weights during training. 
This ensures that the extra branch can simultaneously sustain proper spatial sparsity distribution and re-parameterization capability.
As a result, SpRe can be readily integrated into existing N:M sparsity methods to gain a consistent enhancement in their performance, which is experimentally substantiated in the subsequent section.
We outline the workflow for utilizing SpRe to enhance N:M sparsity methods in Alg.~\ref{alg:spre}.

\begin{table*}[!t]
	\centering
	\caption{Results for sparsifying ResNet-32 at different N:M patterns on CIFAR-10 .}
    \vspace{-0.2cm}
 \renewcommand{\arraystretch}{1.05}
	 \resizebox{0.95\linewidth}{!}{\begin{tabular}[b]{ c|c|cc|cc|cc|cc}
		\toprule
		Model & Method & N:M  & Top-1 Acc & N:M  & Top-1 Acc & N:M  & Top-1 Acc  & N:M  & Top-1 Acc  \\
    \midrule
		ResNet-32 & Baseline & - & {\color{black}95.2$\pm$0.12} & - & {\color{black}95.2$\pm$0.12}& - & {\color{black}95.2$\pm$0.12} & - & {\color{black}95.2$\pm$0.12}\\
		\midrule
		ResNet-32 & ASP~\cite{nvidia2020a100} & 2:4 & {\color{black}95.0$\pm$0.09} & 1:4 & {\color{black}94.3$\pm$0.08} &1:8& {\color{black}93.5$\pm$0.12}& 1:16 & {\color{black}92.6$\pm$0.16}\\
		\gr ResNet-32 &  w. SpRe & 2:4 & \bf {\color{black}95.1$\pm$0.11} & 1:4 & \bf {\color{black}94.7$\pm$0.10} &1:8& \bf {\color{black}94.3$\pm$0.11} & 1:16 & \bf {\color{black}93.7$\pm$0.15}\\
		\midrule
        ResNet-32 & SR-STE~\cite{zhou2021learning} & 2:4 & {\color{black}94.9$\pm$0.19} & 1:4 & {\color{black}94.3$\pm$0.08} &1:8&{\color{black}93.4$\pm$0.14}& 1:16 & {\color{black}92.6$\pm$0.26}\\
		\gr ResNet-32 &  w. SpRe & 2:4 & \bf{\color{black}94.1$\pm$0.17} & 1:4& \bf {\color{black}94.6$\pm$0.12} &1:8& \bf {\color{black}94.3$\pm$0.13}  & 1:16 & \bf {\color{black}93.9$\pm$0.17} \\
        \bottomrule
	\end{tabular}}
    \label{tab:cifar10}
    \vspace{-0.1cm}
\end{table*}

\begin{table*}[!t]
	\centering
	\caption{Results for sparsifying ResNet-18 at different N:M patterns on ImagNet-1K.}
    \vspace{-0.2cm}
 \renewcommand{\arraystretch}{1.05}
 \resizebox{0.95\linewidth}{!}{\begin{tabular}[b]{ c|c|cc|cc|cc|cc}
		\toprule
		Model & Method & N:M  & Top-1 Acc & N:M  & Top-1 Acc & N:M  & Top-1 Acc   & N:M  & Top-1 Acc  \\
    \midrule
		ResNet-18 & Baseline & - & 70.9 & - & 70.9 & - & 70.9 & - & 70.9\\
         \midrule
		ResNet-18 & ASP~\cite{nvidia2020a100} & 2:4 & 70.6 & 1:4 & 69.1 &1:8 & 67.6 & 1:16 & 65.0\\
		\gr ResNet-18 &  w. SpRe & 2:4 &\bf 70.8 & 1:4 & \bf 69.6 &1:8 & \bf 68.2& 1:16 & \bf 65.5 \\
		\midrule
        ResNet-18 & SR-STE~\cite{zhou2021learning} & 2:4 & 71.2 & 1:4 & 69.2 &1:8 & 67.2& 1:16 & 64.9\\
		\gr ResNet-18 &  w. SpRe & 2:4 & \bf 71.3 & 1:4 & \bf 69.9 &1:8 & \bf 68.0& 1:16 & \bf65.7\\
        \bottomrule
	\end{tabular}}
    \label{tab:resnet-18}
    \vspace{-0.1cm}
\end{table*}

\subsection{Tractability of Optimization}

In this section, we theoretically analyze how SpRe compensates spatial sparsity variations of N:M methods. 
Our investigation begins with the BN layer in Eq.~(\ref{eq:forward_inference}), which enables distinct gradient flows across different branches, thereby facilitating multi-branch optimization to enhance spatial sparsity diversity.
Particularly, let $\mathcal{L}$ be the loss function. The gradients for the main branch ($\B \odot \W$) and extra branch ($\B^S \odot \W^S$) are computed as:
\begin{equation}
    \frac{\partial \mathcal{L}}{\partial \W} = \B \odot \left(\frac{\partial \mathcal{L}}{\partial \Y} \circledast \X\right) \odot \gamma_1 \cdot \sigma_1^{-1},
\end{equation}
\begin{equation}
    \frac{\partial \mathcal{L}}{\partial \W^S} = \B^S \odot \left(\frac{\partial \mathcal{L}}{\partial \Y} \circledast \X\right) \odot \gamma_2 \cdot \sigma_2^{-1},
\end{equation}
where $\gamma_1,\sigma_1$ and $\gamma_2,\sigma_2$ are the scale and standard deviation parameters of the respective BN layers.
Here, the BN layers maintain separate statistical profiles ($\gamma,\sigma$) for each branch, with
the scale parameters $\gamma_1,\gamma_2$ evolve independently during training. This creates \textit{asymmetric gradient flows} - weights in spatially important regions (where $\B^S_{:,:,u,v}=1$) receive compounded updates from both branches, while other regions only receive updates from the main branch. In contrast, without BN layers, the forward pass would reduce to:
\begin{equation}
    \Y = (\B \odot \W + \B^S \odot \W^S) \circledast \X.
\end{equation}
In this case, the gradients of the two branches become:
\begin{equation}
    \frac{\partial \mathcal{L}}{\partial \W} = \B \odot \left(\frac{\partial \mathcal{L}}{\partial \Y} \circledast \X\right), \quad
    \frac{\partial \mathcal{L}}{\partial \W^S} = \B^S \odot \left(\frac{\partial \mathcal{L}}{\partial \Y} \circledast \X\right).
\end{equation}
The identical gradient $\partial \mathcal{L}/\partial \Y$ would force both branches to evolve similar update, thereby collapsing spatial diversity and reducing SpRe to conventional N:M sparse training. This theoretically justifies the mechanism of SpRe in preserving unstructured-like spatial sparsity throughout training. 
In summary, the above discussed \textit{BN-Induced Optimization Decoupling} allows the extra branch within SpRe to specialize in compensating spatial sparsity variations via its independent normalization statistics. In this manner, SpRe maintains the unstructured-like spatial sparsity distribution during training, therefore leading to enhanced performance of N:M sparse networks.

\begin{table*}[!t]
	\centering
	\caption{Results for sparsifying ResNet-50 at different N:M patterns on ImagNet-1K.}
    \vspace{-0.2cm}
 \renewcommand{\arraystretch}{1.05}
	 \resizebox{0.95\linewidth}{!}{\begin{tabular}[b]{ c|c|cc|cc|cc|cc}
		\toprule
		Model & Method & N:M  & Top-1 Acc & N:M  & Top-1 Acc & N:M  & Top-1 Acc  & N:M  & Top-1 Acc  \\
   \midrule
		ResNet-50 & Baseline & - & 77.2 & - & 77.2 & -& 77.2 & - & 77.2\\
         \midrule
		ResNet-50 & ASP~\cite{nvidia2020a100} & 2:4 & 77.4 & 1:4 & 76.5 & 1:8 & 75.6 & 1:16 & 71.5 \\
		\gr ResNet-50 &  w. SpRe & 2:4 & \bf 77.7 & 1:4 & \bf 76.8 & 1:8 & \bf 76.2 & 1:16 & \bf 72.3\\
		\midrule
        ResNet-50 & SR-STE~\cite{zhou2021learning} & 2:4 & 77.0 & 1:4 & 75.3  & 1:8 & 73.8& 1:16 & 71.5\\
		\gr ResNet-50 &  w. SpRe & 2:4 & \bf 77.2& 1:4 & \bf76.1 & 1:8 &\bf 74.7& 1:16 & \bf72.7\\
        \midrule
        ResNet-50 & LBC~\cite{zhang2022learning} & 2:4 &77.2 & 1:4 & 75.9  & 1:8 & 74.0& 1:16 & 71.8 \\
		\gr ResNet-50 &  w. SpRe & 2:4 &\bf 77.3 & 1:4 & \bf \bf 76.4 & 1:8 & \bf 74.8  & 1:16 & \bf72.9 \\
        \bottomrule
	\end{tabular}}
    \label{tab:resnet-50}
    \vspace{-0.2cm}
\end{table*}
\begin{table*}[!t]
	\centering
	\caption{Results for sparsifying MobileNet-V1 at different N:M patterns on ImageNet-1K.}
       \vspace{-0.2cm}
 \renewcommand{\arraystretch}{1.05}
	 \resizebox{0.95\linewidth}{!}{\begin{tabular}[b]{ c|c|cc|cc|cc|cc}
		\toprule
		Model & Method & N:M  & Top-1 Acc & N:M & Top-1 Acc & N:M & Top-1 Acc  & N:M  & Top-1 Acc \\
         \midrule
		MobileNet-V1 & Baseline & - & 70.9 & - & 70.9 & - & 70.9& - & 70.9\\
		\midrule
		MobileNet-V1 & ASP~\cite{nvidia2020a100} & 2:4 & 70.2 & 1:4 & 63.9  & 1:8 & 61.4 & 1:16 & 50.4\\
		\gr MobileNet-V1 &  w. SpRe & 2:4 & \bf 70.5& 1:4 & \bf 66.6 & 1:8 & \bf 65.9 & 1:16 & \bf 58.0\\
		\midrule
        MobileNet-V1 & SR-STE~\cite{zhou2021learning} & 2:4 &70.4 & 1:4 & 63.2  & 1:8 & 52.1& 1:16 & 25.4\\
		\gr MobileNet-V1 &  w. SpRe & 2:4 & \bf 70.8 & 1:4 & \bf 64.9 & 1:8 & \bf 54.8 & 1:16 & \bf 34.2 \\
        \bottomrule
	\end{tabular}}
    \vspace{-0.1cm}
    \label{tab:mobv1}
\end{table*}
\begin{table*}[t]
	\centering
	\caption{Results for sparsifying MobileNet-V2 at different N:M patterns on ImageNet-1K.}
    \vspace{-0.2cm}
 \renewcommand{\arraystretch}{1.05}
	\resizebox{0.95\linewidth}{!}{\begin{tabular}[b]{ c|c|cc|cc|cc|cc}
		\toprule
		Model & Method & N:M  & Top-1 Acc & N:M & Top-1 Acc & N:M & Top-1 Acc  & N:M  & Top-1 Acc \\
         \midrule
		MobileNet-V2 & Baseline & - & 72.5 & - & 72.5 & - & 72.5 & - & 72.5\\
		\midrule
		MobileNet-V2 & ASP~\cite{nvidia2020a100} & 2:4 & 71.0& 1:4 &67.4  & 1:8 & 64.0& 1:16 & 58.1\\
		\gr MobileNet-V2 &  w. SpRe & 2:4 & \bf 71.3 & 1:4 & \bf 67.9 & 1:8 & \bf 65.2 & 1:16 & \bf 60.4\\
		\midrule
        MobileNet-V2 & SR-STE~\cite{zhou2021learning} & 2:4 &69.5 & 1:4 & 63.2  & 1:8 & 53.3 & 1:16 & 24.2\\
		\gr MobileNet-V2 &  w. SpRe & 2:4 & \bf 69.9 & 1:4 & \bf 64.3 & 1:8 & \bf 55.0& 1:16 & \bf 35.4 \\
        \bottomrule
	\end{tabular}}
    \label{tab:mobv2}
    \vspace{-0.3cm}
\end{table*}

\begin{algorithm}[!t]
\SetKwInOut{Input}{Require}
\SetKwInOut{Output}{Output}
\caption{Spatial Re-parameterization for N:M Sparsity.}
\label{alg:spre}
\Input{ Network weights $\W$, total training epoch $\tau$.}
$\W^S$ $\gets$ randomly initialization;
\For{$k$ $\gets$ 1, 2, $\dots$, $\tau$ } {
\For{each training step $t$ }  {
Obtain N:M sparse mask $B$; \tcp{Depending on N:M method choosen}
Obtain spatial mask $\B^S$ via Eq.~(\ref{eq:spatial_mask});
Forward and backward propagation via Eq.~(\ref{eq:forward_train});
Update $\W, \W^S$ using SGD optimizer;
}
}
Obtain the merged weights $\W^*$ via Eq.~(\ref{eq:merge}); \\
\noindent \textbf{Output:} Trained N:M sparse weights $\W^*$.
\end{algorithm}

\section{Experiments}\label{experiment}

\subsection{Settings}

Our experiments include three computer vision tasks encompassing image classification on the CIFAR-10~\cite{krizhevsky2009learning} and ImageNet-1K datasets~\cite{deng2009imagenet}, object detection and instance segmentation on the COCO benchmark~\cite{lin2014microsoft}.
The specific experimental settings are respectively expounded as follows.
We validate the efficacy of SpRe in elevating the performance of prominent N:M sparsity methods, including ASP~\cite{nvidia2020a100}, SR-STE~\cite{zhou2021learning}, and LBC~\cite{zhang2022learning}. 
When SpRe is particularly applied to a N:M method, the extra block is initialized and trained with the same configuration as the main block.
Our experiments cover a wide range of N:M patterns including 2:4, 1:4, 1:8, and 1:16.
For image classification, we sparsify ResNet-32~\cite{he2016deep} on CIFAR-10 dataset, and ResNet-18~\cite{he2016deep}, ResNet-50~\cite{he2016deep}, MobileNet-V1~\cite{howard2017mobilenets}, MobileNet-V2~\cite{sandler2018mobilenetv2} on ImageNet-1K dataset.
Besides, we exploit the efficacy of SpRe to aid SR-STE~\cite{zhou2021learning} in training 2:4, 1:4, 1:8, and 1:16 sparse {\color{black}Faster-RCNN~\cite{girshick2015fast}} for object detection and {\color{black}Mask-RCNN~\cite{he2017mask}} for instance segmentation, utilizing ResNet-50 as the backbone.
We employ SpRe to all N:M weights except for 1$\times$ 1 kernels wherein the notion of spatial sparsity is deemed inapplicable.
Following previous works, we train all networks for 300 epochs on CIFAR-10, with a weight decay of 0.005. {\color{black}On ImageNet, 120 epochs are given for ResNet and MobileNet-V1/V2}.
{\color{black}The learning rate is initialized to 0.1 and then decayed using cosine annealing scheduler.
Besides, we follow previous works to use data augmentation including random crop and random horizontal flip~\cite{zhou2021learning}. 
}
For object detection and instance segmentation tasks, we use MMDetection~\cite{chen2019mmdetection} with a training pipeline aligned with SR-STE~\cite{zhou2021learning}.
{\color{black}On CIFAR-10, we conducted five independent trials for each method using different random seeds and report the results as mean $\pm$ standard deviation. For other large-scale datasets, we performed a single experiment due to the stability of the results.
We report the performance of N:M sparsity methods both with and without SpRe, along with the performance of original pre-trained model, denoted as 'Baseline'.
}
Our experiments are implemented with PyTorch~\cite{pytorch2015} and run on NVIDIA Tesla A100 GPUs. 

%

\subsection{Image Classification}
\textbf{CIFAR-10}. We first evaluate the efficacy of SpRe for sparsifying ResNet-32 on the CIFAR-10 dataset, which includes 50,000 training images and 10,000 validation images within 10 classes.
Table\,\ref{tab:cifar10} shows that SpRe can be well leveraged to enhance the performance of all N:M methods.
For instance, the Top-1 classification accuracy of SR-STE is improved by 0.2\%, 0.3\%, 0.9\%, and 1.3\% for 2:4, 1:4, 1:8, and 1:16 patterns, respectively.
Given introducing no extra inference, the effectiveness of SpRe for compensating the limited variation on spatial sparsity of existing methods is obvious.
\noindent \textbf{ImageNet-1K}. For the large-scale ImageNet-1K dataset that contains over 1.2 million images for training and 50,000 images for validation in 1,000 categories, we first present the quantitative results for sparsifying ResNet~\cite{he2016deep} with depths of 18 in Table\,\ref{tab:resnet-18} and 50 in Table\,\ref{tab:resnet-50}.
Encouragingly, SpRe substantially enlarges the performance of N:M methods over a wide range of sparse patterns.
Without extra inference burden introduced, $0.2\%$ and $0.5\%$ Top-1 accuracy improvements are gained by equipping ASP~\cite{nvidia2020a100} with SpRe when sparsifying ResNet-18 at 2:4 and 1:4 patterns.
The advantage of SpRe becomes more evident with an increase in the sparsity rate, where classic N:M sparsity methods fail to allocate enough weights for processing the important visual points due to a constant spatial sparsity.
Upon training 1:16 sparse ResNet-50, SpRe exhibits a remarkable enhancement in the Top-1 accuracy of SR-STE~\cite{zhou2021learning} and LBC~\cite{zhang2022learning} by 1.2\% and 1.1\%, respectively. 
The improvement is attributed to our weight preservation at crucial visual locations {\color{black}at each local receptive field}.
Therefore, the efficacy of SpRe on the large-scale challenging tasks is also evident.

\begin{figure}[!t]
\begin{center}
\includegraphics[width=0.95\linewidth]{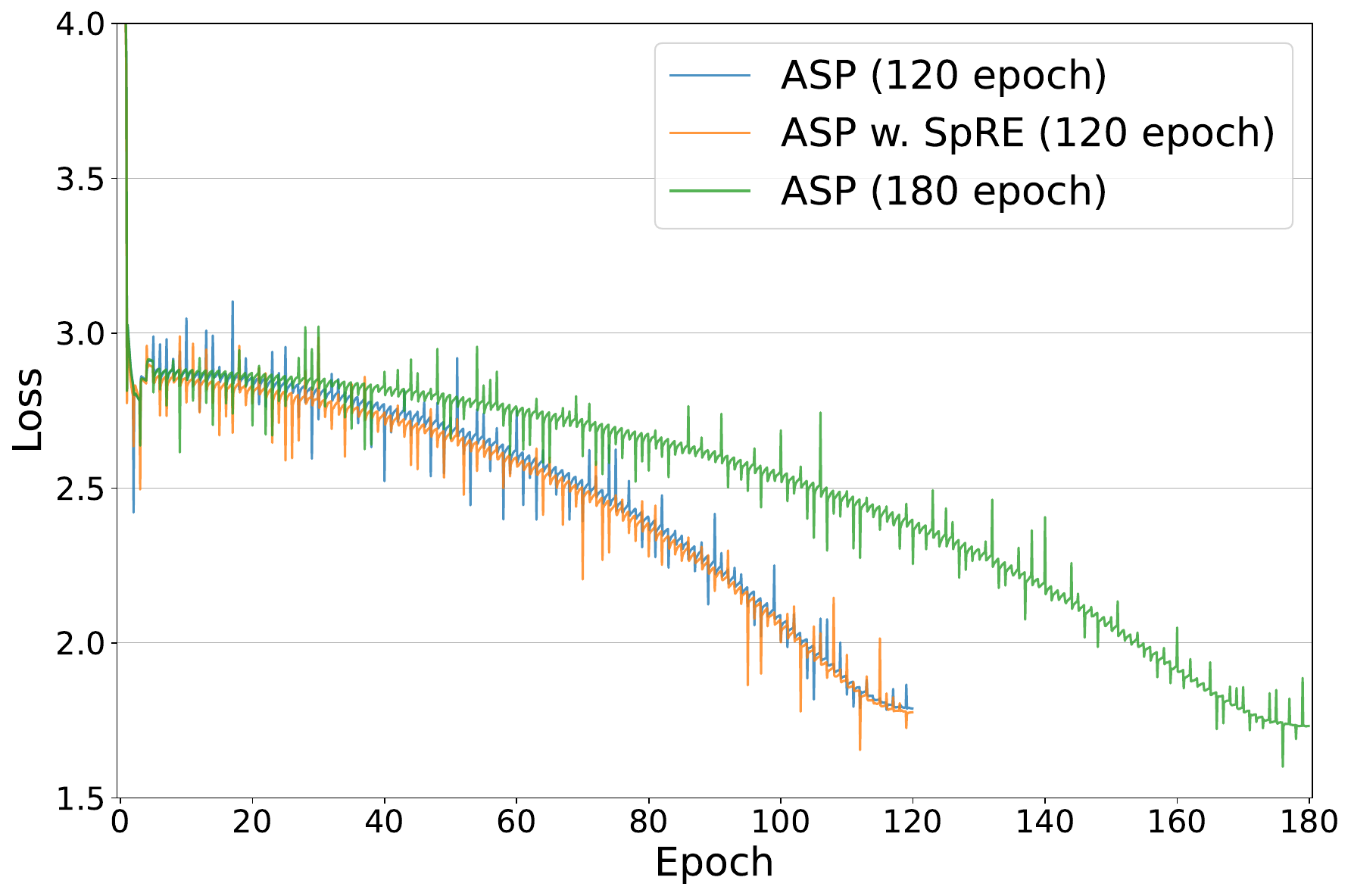}
\end{center}
\vspace{-0.4cm}
{\color{black}\caption{\label{fig:loss_curve}Loss curve during training 1:16 sparse ResNet-50 using different methods on ImageNet-1k. The final Top-1 accuracy are 72.3\%, 71.5\%, and 71.8\% for ASP w. SpRe (120 epoch), ASP (120 epoch), and ASP (180 epoch), respectively. 
}}
\vspace{-0.5cm}
\end{figure}

\begin{table}[!t]
	\centering
	\caption{Performance comparison between N:M and unstructured sparsity methods for sparsifying ResNet-50 on ImageNet-1K. $\dag$ denotes our reproduced results with 120 training epochs given.}
    \vspace{-0.2cm}
     \renewcommand{\arraystretch}{0.98}
	\resizebox{0.45\textwidth}{!}{\begin{tabular}[b]{ l|cc|c }
	\toprule
	 Method & Sparsity & Top-1 Acc &Structured  \\
  \midrule
		ResNet50 & 0 &   77.3 &-  \\
  \midrule
	 DNW~\cite{wortsman2019discovering} &  90    &74.0     &\xmark \\
 RigL~\cite{evci2020rigging}   & 90 & 73.0  &\xmark \\
{\color{black}RigL$^\dag$~\cite{evci2020rigging}}   & {\color{black}90} & {\color{black}73.4}  &{\color{black}\xmark} \\
    STR~\cite{kusupati2020soft} &  91 & 74.0    &\xmark \\
      GraNet~\cite{liu2021sparse}  &  90 & 74.5  &\xmark \\
      {\color{black}GraNet$^\dag$~\cite{liu2021sparse}}  &  {\color{black}90} & {\color{black}74.7}  &{\color{black}\xmark} \\
    \midrule
	  SR-STE~\cite{zhou2021learning} &  88(1:8) & 73.8 &\cmark\\
	\gr  	\bf  w. SpRe& 88(1:8) & \bf 74.7&\cmark\\
 \midrule
       LBC~\cite{zhang2022learning} &88(1:8) &  74.0 &\cmark \\
	\gr \bf  w. SpRe & 88(1:8) & \bf 74.8 &\cmark \\
 \midrule
 	  DNW~\cite{wortsman2019discovering}   &  95    &68.3 &\xmark \\
	  RigL~\cite{evci2020rigging}   & 95 & 70.0  &\xmark \\
     {\color{black} RigL$^\dag$~\cite{evci2020rigging}}   & {\color{black}95} & {\color{black}70.3}  &{\color{black}\xmark} \\
    STR~\cite{kusupati2020soft}  &  95 & 70.4 &\xmark \\
      GraNet~\cite{liu2021sparse}   &  95 & 72.3 &\xmark \\
      {\color{black}GraNet$^\dag$~\cite{liu2021sparse}}   &  {\color{black}95} & {\color{black}72.9} &{\color{black}\xmark} \\
    \midrule
	  SR-STE~\cite{zhou2021learning} & 94(1:16) & 71.5  &\cmark\\
	\gr  	\bf  w. SpRe& 94(1:16) & \bf 72.7 &\cmark\\
 \midrule
       LBC~\cite{zhang2022learning}& 94(1:16) & 71.8 &\cmark \\
	\gr  \bf  w. SpRe & 94(1:16) & \bf 72.9&\cmark \\
        \bottomrule
	\end{tabular}}
    \label{tab:com_with_unstruc}
    \vspace{-0.0cm}
\end{table}

Furthermore, we examine the generalization capability of SpRe for sparsifying MobileNet-V1~\cite{howard2017mobilenets} and MobileNet-V2, two lightweight networks incorporating depth-wise convolution design and posing more challenges for compression.
Table\,\ref{tab:mobv1} and Table\,\ref{tab:mobv2} respectively show the quantitative results.
It can be seen that all N:M methods suffer significantly more performance drops for sparsifying lightweight networks. 
Nevertheless, SpRe demonstrates a distinct trend whereby the performance improvements remain consistent and increase as the sparsity level increases, irrespective of the specific N:M methods employed.
For instance, SpRe is able to enhance the Top-1 accuracy of ASP by $0.3\%$, $2.5\%$, $4.5\%$and $7.6\%$ when sparsifying MobileNet-V1 at 2:4, 1:4, 1:18, 1:16 sparse patterns, respectively.
Similar phenomenon can be drawn when it comes to MobileNet-V2.
These results well highlight the efficacy of SpRe in advancing existing N:M sparsity methods for sparsifying lightweight networks.

\begin{table}[!t]
\centering
\caption{Results for object detection on COCO.}
\label{tab:object}
\vspace{-0.2cm}
\resizebox{0.71\linewidth}{!}{\begin{tabular}{cccc}
\toprule
Model & Method  & N:M & \quad mAP \quad\\
\midrule
F-RCNN & Baseline & - & 37.4\\
\midrule
F-RCNN & SR-STE & 2:4 & 38.2\\
\gr F-RCNN & \bf  w. SpRe & 2:4 & \bf 38.4\\
\midrule
F-RCNN & SR-STE & 1:4 & 37.1\\
\gr F-RCNN & \bf  w. SpRe & 1:4 & \bf 37.3\\
\midrule
F-RCNN & SR-STE & 1:16 & 35.8\\
\gr F-RCNN & \bf  w. SpRe & 1:16 & \bf 36.3\\
\bottomrule
\end{tabular}}
\vspace{-0.4cm}
\end{table}

{\color{black}Additionally, we examine the effect of SpRe on the convergence speed of N:M sparsity methods. Specifically, Figure \ref{fig:loss_curve} illustrates the loss curves of ASP and ASP w.SpRe when training a 2:4 sparse ResNet-50. Additionally, we conduct ablation experiments by extending the training epochs to 180 to ensure a comprehensive comparison under full convergence. The results demonstrate that SpRe consistently outperforms in both loss convergence speed and final model accuracy, enhancing the convergence efficiency of N:M sparsity methods.}

At last, we provide the performance comparison between unstructured sparsity and N:M sparsity methods in Table\,\ref{tab:com_with_unstruc}.
The results reveal that N:M sparsity methods struggle to maintain accuracy on par with advanced unstructured sparsity techniques at comparable levels of sparsity.
This disparity in performance can be attributed to the fact that unstructured sparsity, as we previously discussed, sustains adequate processing of crucial visual points at high levels of sparsity. 
In contrast, N:M sparsity fails to effectively handle this aspect due to its constant spatial sparsity and hence lags behind in achieving comparable outcomes.
Fortunately, by allocating re-parameterizable weights that follow the spatial sparsity distribution of unstructured sparsity, SpRe effectively elevates the accuracy of N:M sparsity methods to the level of state-of-the-art unstructured sparsity methods without incurring any extra inference overhead.
For instance, SR-STE surpasses the recent unstructured sparsity method GraNet~\cite{liu2021sparse} by 0.4\% Top-1 accuracy with the aid of SpRe (72.7\% for SR-STE boosted by SpRe and 72.3\% for GraNet at similar sparsity rates).
Given the distinct advantage of N:M sparsity for practical acceleration on the N:M sparse tensor core, the significance of SpRe in bridging the performance gap between N:M sparsity and unstructured sparsity is apparent.

\begin{table}[!t]
 \centering
\caption{Results for instance segmentation on COCO.}\label{tab:segment}
\vspace{-0.2cm}
\resizebox{0.95\linewidth}{!}{\begin{tabular}{ccccc}
\toprule
Model & Method  & N:M & Box mAP & Mask mAP\\
\midrule
M-RCNN & Baseline & - &38.2	& 34.7\\
\midrule
M-RCNN & SR-STE & 2:4 & 	39.0&	35.3\\
\gr M-RCNN & \bf  w. SpRe & 2:4 & \bf 39.2 &	\bf 35.7\\
\midrule
M-RCNN & SR-STE & 1:4 & 37.4&	33.5\\
\gr M-RCNN & \bf  w. SpRe & 1:4 & \bf 37.7 &	\bf 33.9\\
\midrule
M-RCNN & SR-STE & 1:16 & 35.4&	31.8\\
\gr M-RCNN & \bf  w. SpRe & 1:16 & \bf 36.1 &	\bf 32.5\\
 \bottomrule
\end{tabular}}
\vspace{-0.2cm}
\end{table}

\begin{figure}[!t]
\begin{center}
\includegraphics[width=0.95\linewidth]{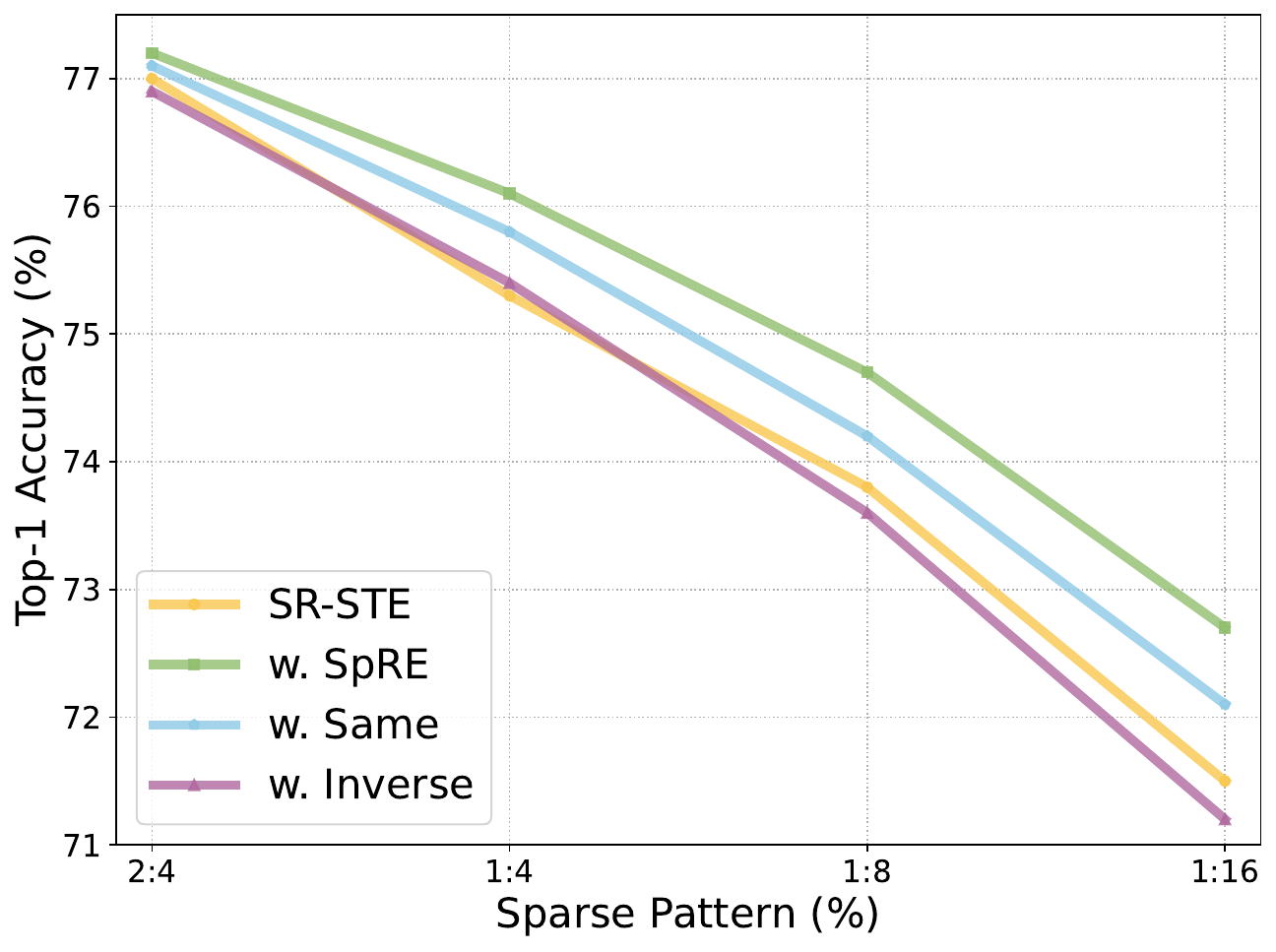}
\end{center}
\vspace{-0.3cm}
\caption{
Performance comparison for using different strategies to generate the extra branch. Experiments are based on using SR-STE to sparsify ResNet-50 on ImageNet-1k.
}\label{fig:extra_branch}
\vspace{-0.2cm}
\end{figure}
%


\subsection{Object Detection and Instance Segmentation}
Beyond fundamental image classification benchmarks, we exploit the generalization ability of SpRe on the object detection and instance segmentation tasks of COCO benchmark~\cite{lin2014microsoft}.
Table\,\ref{tab:object} compares our proposed SpRe to SR-STE for training N:M sparse Faster-RCNN~\cite{girshick2015fast}.
Notably, SpRe yields robust performance improvement of $0.2$, $0.2$ and $0.5$ mAP at 2:4, 1:4, and 1:16 sparse patterns, respectively.
Similar trends can be observed from Table\,\ref{tab:segment} when sparsifying Mask-RCNN~\cite{he2017mask} for the instance segmentation task. 
These results well substantiate the robustness and effectiveness of SpRe on downstream computer vision tasks.

\subsection{Performance Analysis}
{\textbf{Extra Branch.}} We first present the performance analysis of SpRe by investigating two variants on the derivation of the extra branch in Eq.\,(\ref{eq:spatial_mask}).
The experiments incorporate 2:4, 1:4, and 1:16 patterns to sparsify ResNet-50 on ImageNet-1K, with SR-STE~\cite{zhou2021learning} serving as the comparative baseline. 
Examining the details, we first perform ablation that maintains the same mask with the main N:M sparse weights as $\B^S = \B$.
This variation carries out more preserved weights in the extra branch, despite the spatial sparsity mirroring that of the vanilla N:M sparsity.
As depicted in Fig~\ref{fig:extra_branch}, more weights in the extra branch (referred to as Same) even bring less performance improvement.
We attribute this phenomenon to limited variability in spatial sparsity as discussed in Sec.~\ref{sec:spatial_vari}.
Besides, we consider another variation that allocates parameters at spatial locations where unstructured sparsity exhibits less spatial sparsity than N:M sparsity (referred to as Inverse).
As can be seen, the inverse allocation for the weights in the extra branch brings even negative performance, which further demonstrates our point that a correct spatial sparsity variation is the key to the performance retention of sparse networks.

\begin{table}[!t]
\caption{{\color{black}Theoretical} FLOPs and parameters comparison for pruning ResNet-50 on ImageNet-1K at training and testing stages.}
\vspace{-0.2cm}
\tabcolsep=5.0pt
\centering
\label{tab:cost}
\resizebox{0.95\linewidth}{!}{\begin{tabular}{lccccccc}
\toprule
 Method  & N:M & FLOPs	&FLOPs&	Params&	Params\\
 & & (Train) & (Test) & (Train) & (Test) \\
\midrule
ResNet-50 & - &1$\times$&	1$\times$	&1$\times$	&1$\times$ \\
\midrule
SR-STE & 2:4 &  0.83$\times$	&0.5$\times$	&1$\times$	&0.5$\times$\\
w. SpRe & 2:4 & 1.02$\times$ &	0.5$\times$	&1.24$\times$&	0.5$\times$\\
\midrule
SR-STE & 1:4 & 0.74$\times$ &	0.25$\times$	&1$\times$	&0.25$\times$ \\
w. SpRe & 1:4 &0.80$\times$&	0.25$\times$&	1.14$\times$	&0.25$\times$ \\
\midrule
SR-STE & 1:8 & 0.71$\times$ & 0.13$\times$ & 1$\times$ & 0.13$\times$\\
w. SpRe & 1:8 &  0.73$\times$ & 0.13$\times$ & 1.09$\times$ & 0.13$\times$\\
\midrule
SR-STE  & 1:16 & 0.69$\times$ &	0.06$\times$&	1$\times$	&0.06$\times$ \\
w. SpRe & 1:16 &  0.70$\times$ &	0.06$\times$	&1.05$\times$	&0.06$\times$\\
\bottomrule
\end{tabular}}
\end{table}

\begin{table}[!t]
\tabcolsep=5.0pt
\centering
{\color{black}\caption{\label{tab:cost2}Training latency (A100 GPU hour/epoch) and Memory cost (GB) for pruning ResNet-50.}}
\vspace{-0.2cm}
\resizebox{0.7\linewidth}{!}{\begin{tabular}{lccccc}
\toprule
 {\color{black}Method}  & {\color{black}N:M} & {\color{black}Training}	& {\color{black}Memory}\\
 & & {\color{black}Latency} &  {\color{black}Cost} \\
\midrule
{\color{black}ResNet-50 }& {\color{black}- }&{\color{black} 0.39 }& {\color{black}27.9G}\\
\midrule
{\color{black}ASP} & {\color{black}2:4} & {\color{black}0.40} & {\color{black}28.0G}\\
{\color{black}w. SpRe} & {\color{black}2:4} & {\color{black}0.42} & {\color{black}28.3G}\\
\midrule
{\color{black}SR-STE} & {\color{black}2:4} & {\color{black}0.41} & {\color{black}28.2G} \\
{\color{black}w. SpRe} & {\color{black}2:4} &{\color{black}0.44} & {\color{black}28.6G}\\
\bottomrule
\end{tabular}}
\vspace{-0.3cm}
\end{table}

\noindent {\textbf{Additional Overhead.}} We further elucidate the overhead implicated in Floating Point Operations (FLOPs) and parameters concomitant with the introduction of SpRe.
Table\,\ref{tab:cost} enumerates the {\color{black}theoretical} results from employing SR-STE for the sparsification of ResNet-50 on ImageNet-1K at different N:M patterns.
{\color{black}In addition, Table\,\ref{tab:cost2} report the training memory and speed results from employing ASP and SR-STE at 2:4 sparsity.}
The deployment of SpRe does, indeed, incur a modicum of additional overhead during the training phase, albeit minimal.
However, during the inference (test) stage, the quantity of parameters and FLOPs perseveres consistently, enabling efficacious deployment and acceleration founded on the Sparse Tensor Core~\cite{nvidia2020a100}. 
Concurrently, juxtaposed with the results presented in Table\,\ref{tab:cost}, SpRe achieves a palpable enhancement in performance with the introduction of merely a negligible training overhead, affirming its undoubted practical utility.

\section{Limitation}

We further discuss unexplored limitations, which will be our future focus. 
SpRe is particularly presented for N:M sparsity in convolutional neural networks upon our observations of spatial sparsity variability.
{\color{black}
Although not the focus of this current work, it would be interesting for future work to examine similar discoveries to drive further enhancement for N:M sparsity in networks with different typologies,~\emph{e.g.}, Vision Transformers (ViTs), BERT.
While non-CNN architectures do not inherently incorporate spatial sparsity, investigating the distributional disparities between unstructured and N:M sparsity patterns, and further designing corresponding re-parameterization mechanisms presents a promising approach to improve the performance of N:M sparse networks.
}
Besides, though improving the performance, SpRe also brings some extra burdens for N:M sparsity in the training time.
A promising direction in the future is to develop more efficient ways to mitigate the lack in spatial sparsity variability of N:M sparsity.

\section{Conclusion}
In this work, we have presented Spatial Re-parameterization, an effective and easy-to-use method for N:M sparsity.
By introducing a re-parameterizable branch that follows the spatial sparsity distribution of unstructured sparsity, SpRe is able to reimburse the spatial sparsity variability during the training time of N:M sparsity, which we examine to be the core for performance retention.
Our proposed SpRe is benchmarked on several computer vision benchmarks, consistently delivering enhanced performance for representative N:M methods without extra inference burden.
Notably, SpRe brings the performance of N:M sparsity methods to a comparable level with unstructured sparsity methods for the first time.
Hopefully, this work shields a convincing trail to dive into the intrinsic property of N:M sparsity.

\section*{Acknowledgement}
This work was supported by National Key R\&D Program of China (No.2022ZD0118202), the National Science Fund for Distinguished Young Scholars (No.62025603), the National Natural Science Foundation of China (No.624B2119, No. U21B2037, No. U22B2051, No. 62176222, No. 62176223, No. 62176226, No. 62072386, No. 62072387, No. 62072389, No. 62002305 and No. 62272401), and the Natural Science Foundation of Fujian Province of China (No.2021J01002,  No.2022J06001).

\bibliographystyle{IEEEtran}
\bibliography{main}


%



\ifCLASSOPTIONcaptionsoff
  \newpage
\fi

\begin{IEEEbiography}[{\includegraphics[width=1in,height=1.25in,clip,keepaspectratio]{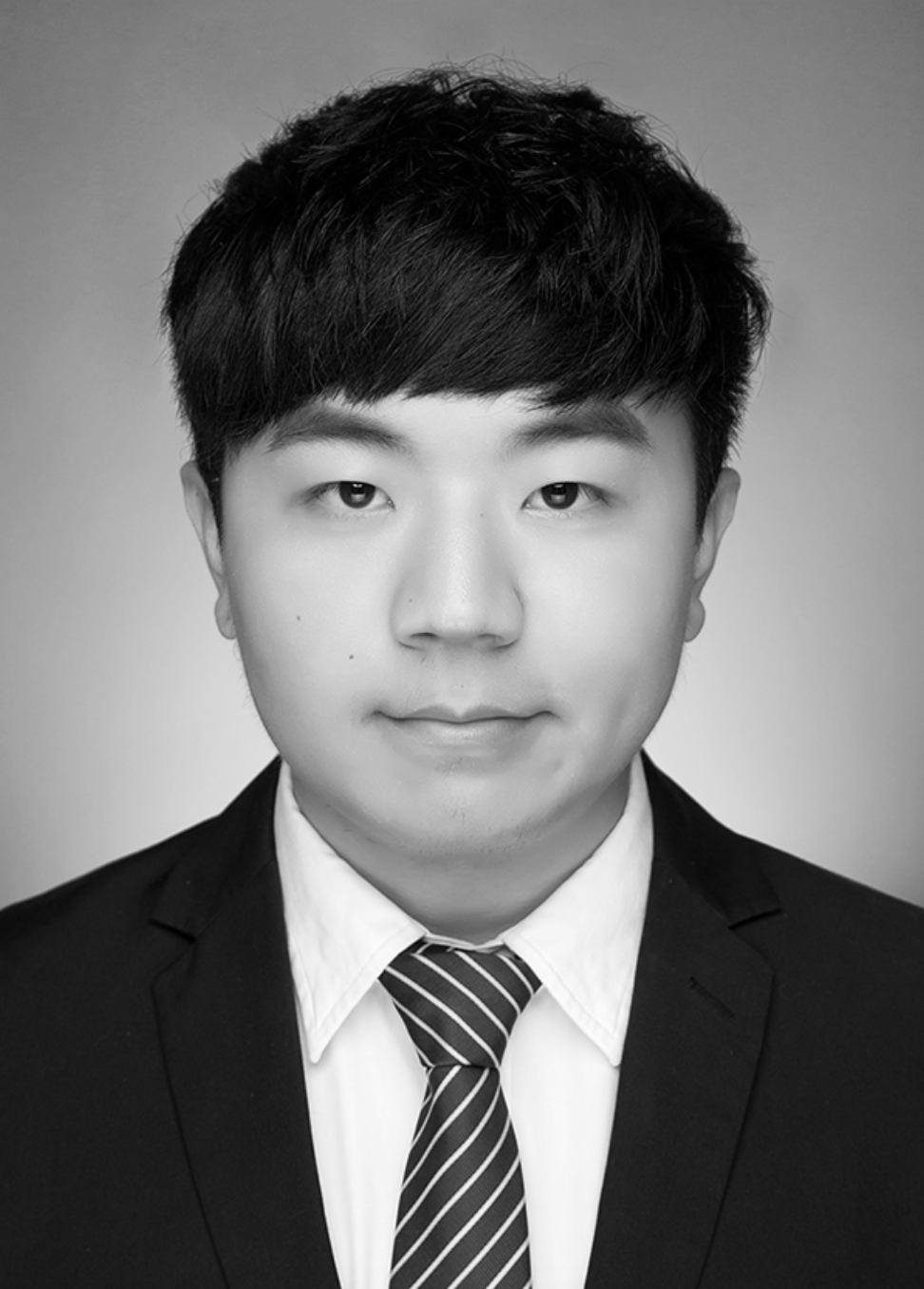}}]{Yuxin Zhang} received the B.E. degree in Computer Science, School of Informatics, Xiamen University, Xiamen, China, in 2020.
He is currently pursuing the P.H.D degree with Xiamen University, China. His publications on top-tier conferences/journals include IEEE TPAMI, IEEE TNNLS, NeurIPS, ICLR, ICML, CVPR, ICCV, IJCAI and so on. His research interests include computer vision and neural network compression \& acceleration.
\end{IEEEbiography}

\begin{IEEEbiography}[{\includegraphics[width=1in,height=1.25in,clip,keepaspectratio]{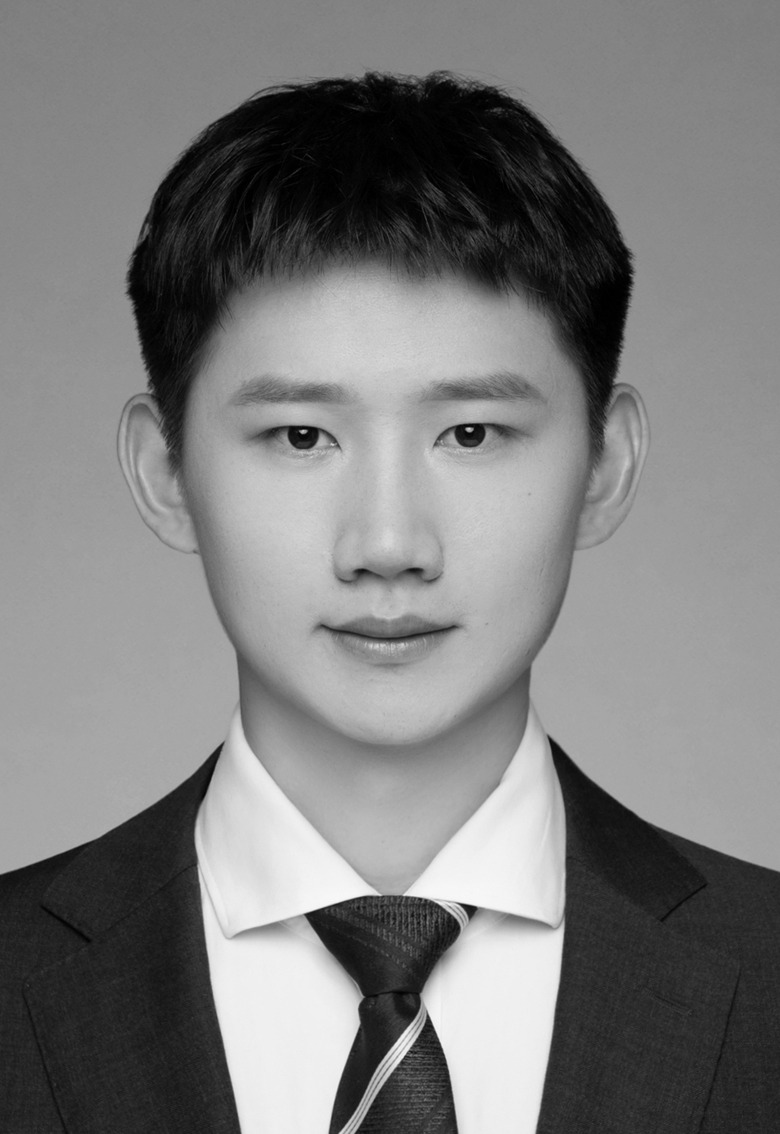}}]{Mingbao Lin} finished his M.S.-Ph.D. study and obtained the Ph.D. degree in intelligence science and technology from Xiamen University, Xiamen, China, in 2022. Earlier, he received the B.S. degree from Fuzhou University, Fuzhou, China, in 2016.

He is currently a researcher with Skywork AI, Singapore. His publications on top-tier conferences/journals include IEEE TPAMI, IJCV, IEEE TIP, IEEE TNNLS, CVPR, NeurIPS, AAAI, IJCAI, ACM MM and so on. His current research interest is to develop efficient vision model, as well as information retrieval.
\end{IEEEbiography}

\begin{IEEEbiography}[{\includegraphics[width=1in,height=1.25in,clip,keepaspectratio]{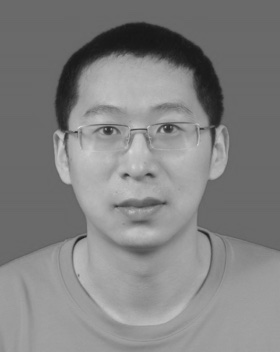}}]{Mingliang Xu} (Member, IEEE) is a professor and dean of the School of Computer and Artificial Intelligence at Zhengzhou University, and the director of the Engineering Research Center of Ministry of Education on Intelligent Swarm Systems, China. He received his Ph.D. degree from the State Key Lab of CAD\&CG at Zhejiang University, Hangzhou, China, and the B.S. and M.S. degrees from the Computer Science Department, Zhengzhou University, Zhengzhou, China, respectively. He was awarded as the National Science Foundation for Distinguished Young Scholars (2023). His research interests include computer graphics, multimedia, and artificial intelligence. He has authored more than 60 journal and conference papers in the above areas, including the ACM Transactions on Graphics, the ACM Transactions on Intelligent Systems and Technology, the IEEE TRANSACTIONS ON PATTERN ANALYSIS AND MACHINE INTELLIGENCE, the IEEE TRANSACTIONS ON IMAGE PROCESSING, the IEEE TRANSACTIONS ON CYBERNETICS, the IEEE TRANSACTIONS ON CIRCUITS AND SYSTEMS FOR VIDEO TECHNOLOGY, ACM SIGGRAPH (Asia), ACM MM, and ICCV model, as well as information retrieval.
\end{IEEEbiography}

\begin{IEEEbiography}[{\includegraphics[width=1in,height=1.25in,clip,keepaspectratio]{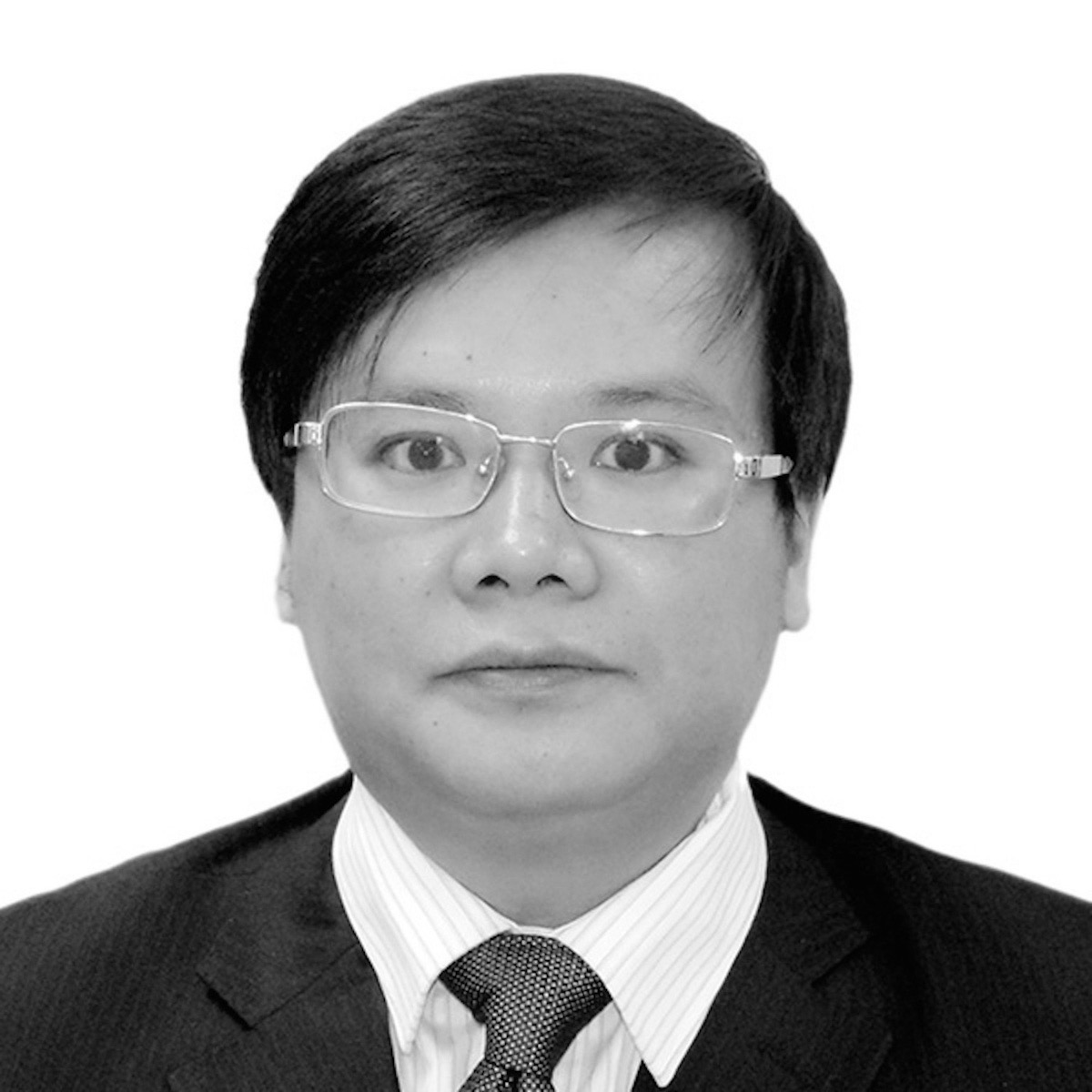}}]{Yonghong Tian} (Fellow, IEEE)
is currently a Boya Distinguished Professor with the Department of Computer Science and Technology, Peking University, China. His research interests include neuromorphic vision, brain-inspired computation and multimedia big data. He is the author or coauthor of over 200 technical articles in refereed journals such as IEEE TPAMI/TNNLS/TIP/TMM/TCSVT/TKDE/TPDS, ACM CSUR/TOIS/TOMM and conferences such as NeurIPS/CVPR/ICCV/AAAI/ACMMM/WWW. Prof. Tian was/is an Associate Editor of IEEE TCSVT (2018.1-), IEEE TMM (2014.8-2018.8), IEEE Multimedia Mag. (2018.1-), and IEEE Access (2017.1-). He co-initiated IEEE Int’l Conf. on Multimedia Big Data (BigMM) and served as the TPC Co-chair of BigMM 2015, and aslo served as the Technical Program Co-chair of IEEE ICME 2015, IEEE ISM 2015 and IEEE MIPR 2018/2019, and General Co-chair of IEEE MIPR 2020 and ICME2021. He is the steering member of IEEE ICME (2018-) and IEEE BigMM (2015-), and is a TPC Member of more than ten conferences such as CVPR, ICCV, ACM KDD, AAAI, ACM MM and ECCV. He was the recipient of the Chinese National Science Foundation for Distinguished Young Scholars in 2018, two National Science and Technology Awards and three ministerial-level awards in China, and obtained the 2015 EURASIP Best Paper Award for Journal on Image and Video Processing, and the best paper award of IEEE BigMM 2018. He is a senior member of IEEE, CIE and CCF, a member of ACM.
\end{IEEEbiography}

\begin{IEEEbiography}[{\includegraphics[width=1in,height=1.25in,clip,keepaspectratio]{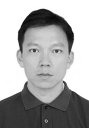}}]{Rongrong Ji}
(Senior Member, IEEE) is a Nanqiang Distinguished Professor at Xiamen University, the Deputy Director of the Office of Science and Technology at Xiamen University, and the Director of Media Analytics and Computing Lab. He was awarded as the National Science Foundation for Excellent Young Scholars (2014), the National Ten Thousand Plan for Young Top Talents (2017), and the National Science Foundation for Distinguished Young Scholars (2020). His research falls in the field of computer vision, multimedia analysis, and machine learning. He has published 50+ papers in ACM/IEEE Transactions, including TPAMI and IJCV, and 100+ full papers on top-tier conferences, such as CVPR and NeurIPS. His publications have got over 10K citations in Google Scholar. He was the recipient of the Best Paper Award of ACM Multimedia 2011. He has served as Area Chairs in top-tier conferences such as CVPR and ACM Multimedia. He is also an Advisory Member for Artificial Intelligence Construction in the Electronic Information Education Committee of the National Ministry of Education.
\end{IEEEbiography}




\end{document}